\documentclass{article} 
\usepackage{iclr2027_conference,times}

\usepackage{amsmath,amsfonts,bm}

\def\eqref#1{equation~\ref{#1}}

\def\1{\bm{1}}

\DeclareMathAlphabet{\mathsfit}{\encodingdefault}{\sfdefault}{m}{sl}
\SetMathAlphabet{\mathsfit}{bold}{\encodingdefault}{\sfdefault}{bx}{n}

\usepackage{hyperref}
\usepackage{url}
\usepackage[pdftex]{graphicx}
\usepackage[accsupp]{axessibility}  
\usepackage[utf8]{inputenc} 
\usepackage[T1]{fontenc}    
\usepackage{url}            
\usepackage{booktabs}       
\usepackage{amsfonts}       
\usepackage{amsthm}
\usepackage{nicefrac}       
\usepackage{microtype}      
\usepackage{xcolor}         

\usepackage{microtype}
\usepackage{subcaption}
\usepackage{pifont}
\usepackage{bm}
\usepackage{wrapfig}
\usepackage{multicol}
\usepackage{multirow}
\usepackage{comment}
\usepackage{enumitem}
\usepackage{marvosym}

\title{RoAR-VAE: Improving Latent Diffusability through Robust Prediction}

\author{%
\parbox{\textwidth}{%
\centering
\vspace{2.0em}
Yang Yue$^{1}$\quad
Fangyun Wei$^{2}$\quad
Yitong Wang$^{2}$\quad
Jinjing Zhao$^{2}$\\[0.2em]
Henry Zheng$^{1}$ \quad
Dong Chen$^{2}$\thanks{Corresponding authors.\quad~Code is available at \href{https://github.com/LeapLabTHU/RoAR-VAE}{\texttt{https://github.com/LeapLabTHU/RoAR-VAE}}.} \quad
Gao Huang$^{1}$\footnotemark[1]\\[0.7em]
\normalfont
$^{1}$Tsinghua University\quad
$^{2}$Microsoft Research\\[0.7em]
{\footnotesize
\mbox{\texttt{yueyang22@mails.tsinghua.edu.cn}}\quad
\mbox{\texttt{doch@microsoft.com}}\quad
\mbox{\texttt{gaohuang@tsinghua.edu.cn}\quad
}
\\[0.7em]
}
}
}

\iclrarxiv
\begin{document}

\maketitle

\begin{abstract}
An effective visual tokenizer must preserve image fidelity while producing latents that downstream diffusion models can learn efficiently.
Prior work improves latent diffusability through semantic alignment or joint tokenizer--diffusion training.
Revisiting REPA-E, a representative approach, we find that removing its diffusion objective and dedicated generative blocks while retaining semantic prediction from corrupted latents yields comparable downstream generation performance under a matched setting. This observation motivates a general principle of \emph{robust prediction}: generation-relevant information should remain recoverable from corrupted latents.
We instantiate this principle with two complementary objectives. \textit{RobustAlign} predicts pretrained semantic features from corrupted latents to preserve high-level structure, while \textit{RobustRecon} reconstructs the clean image to retain fine-grained spatial and appearance information. Together, they form \textbf{Ro}bust \textbf{A}lignment and \textbf{R}econstruction VAE (\textbf{RoAR-VAE}).
Our analysis suggests that robust prediction improves diffusability by making useful information more accessible in high-variance latent directions, which diffusion models tend to learn earlier.
Empirically, RoAR-VAE improves average generation benchmark performance by 2.19 points over the strongest baseline tokenizer under the same training budget, while improving small-text reconstruction accuracy by 16.93 percentage points over FLUX.1-VAE. These results demonstrate that RoAR-VAE achieves a favorable balance between reconstruction fidelity and downstream generative learnability.
\end{abstract}
\section{Introduction}


Visual tokenizers define the representation on which latent diffusion models learn to generate images~\citep{ldm,sd3}. An effective tokenizer should preserve sufficient image information for faithful reconstruction, while organizing that information into a representation that a downstream diffusion model can learn efficiently. This latter property, often referred to as \textit{diffusability}, can conflict with reconstruction fidelity: higher-capacity or weakly regularized latents may retain more visual detail yet be harder to model. Balancing reconstruction quality and diffusability is therefore a central challenge in visual tokenizer design.

\begin{figure}[t]
    \centering
    \includegraphics[width=\linewidth]{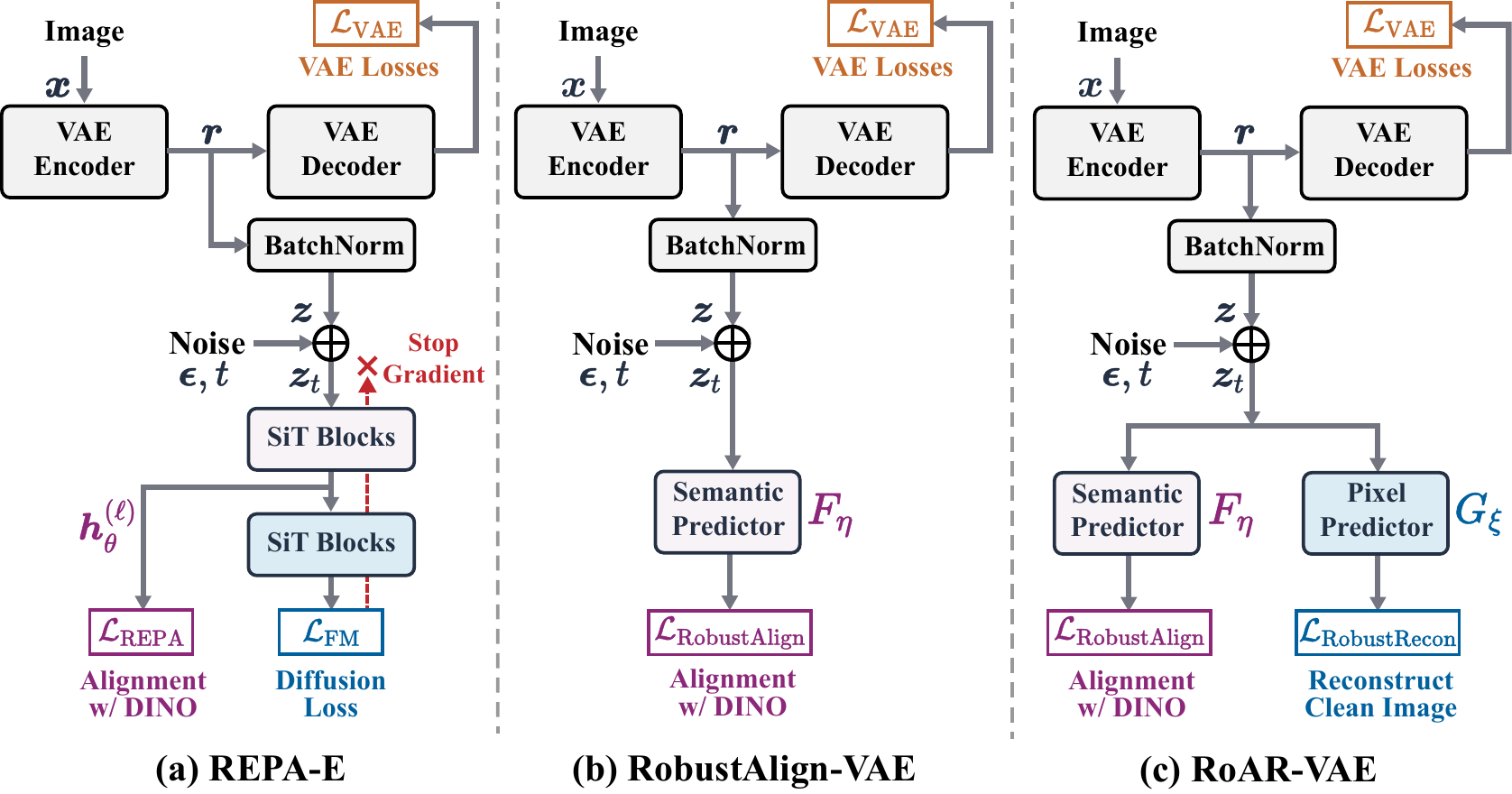}
    \caption{\textbf{From REPA-E to RoAR.} (a) REPA-E jointly trains the tokenizer and diffusion model, but only semantic alignment backpropagates through corrupted latents to the tokenizer encoder. (b) RobustAlign-VAE isolates this semantic-prediction pathway by removing the diffusion objective and its associated modules. (c) RoAR further introduces RobustRecon, combining semantic and image reconstruction targets under latent corruption alongside the standard VAE objective on clean latents.}
    \label{fig:framework}
\end{figure}

Recent work improves latent diffusability through semantic supervision during VAE training~\citep{vavae,vtp} or joint optimization of the tokenizer and diffusion model~\citep{heek2026unified,unite}.
A representative approach, REPA-E~\citep{repae}, combines both ideas: as shown in Fig.~\ref{fig:framework}(a), it jointly trains a VAE and a diffusion Transformer while aligning intermediate features with semantic targets from pretrained visual encoders~\citep{repa,dinov2}.
However, only the semantic alignment objective directly updates the tokenizer encoder, while gradients from the diffusion objective are detached.
This raises a natural question: \emph{what actually drives the diffusability improvement in REPA-E, joint diffusion training or semantic supervision?}  To isolate these factors, we simplify REPA-E by removing the diffusion objective and its dedicated generative blocks while retaining prediction of pretrained visual features from corrupted latents, yielding \emph{RobustAlign} (Figure~\ref{fig:framework}(b)).
Under a matched evaluation, the resulting tokenizer achieves comparable downstream generation performance to the tokenizer trained with REPA-E (Table~\ref{tab:repae}), indicating that noisy semantic prediction alone can provide an effective learning signal for shaping a diffusion-friendly latent space.

RobustAlign suggests a broader principle of \emph{robust prediction}: generation-relevant information should remain predictable from heavily corrupted latents. By backpropagating this prediction objective through the corrupted latent, the tokenizer encoder is encouraged to organize target-relevant information so that it remains accessible under corruption. RobustAlign primarily supervises object-level structure and high-level visual content captured by pretrained features, but may underrepresent fine-grained spatial and appearance variations. We therefore introduce \textbf{RobustRecon}, which applies the same principle to the clean image itself, encouraging detailed image content to remain recoverable from corrupted latents.
Combining the two objectives yields \textbf{\textbf{Robust Alignment and Reconstruction (RoAR)}} (Figure~\ref{fig:framework}(c)).  Both auxiliary objectives shape the encoder through corrupted latents, while the standard VAE reconstruction objective preserves faithful reconstruction from clean latents. 
The latent PCA visualizations in Figure~\ref{fig:pca} further illustrate their complementary effects: RobustAlign emphasizes object-level organization and foreground--background separation, RobustRecon exposes richer local variation, and their combination exhibits both forms of structure.

We further ask: \textit{why does robust prediction improve diffusability?} Under Gaussian corruption, higher-variance latent directions retain a higher signal-to-noise ratio and are therefore more robust to noise. Training the encoder to recover clean targets from corrupted latents consequently encourages generation-relevant information to remain accessible in these dominant directions, potentially by strengthening the principal modes or by enriching the image information they encode, as suggested by the PCA visualizations in Fig.~\ref{fig:pca}. Prior analyses further show that higher-variance covariance modes tend to be learned earlier by diffusion models~\citep{wang2025an}. Together, these observations suggest that robust prediction organizes useful information into prominent latent structure that can be acquired more readily by a downstream diffusion model.

\begin{wrapfigure}{r}{0.55\textwidth}
\vspace{-15pt}
\centering
\includegraphics[width=\linewidth]{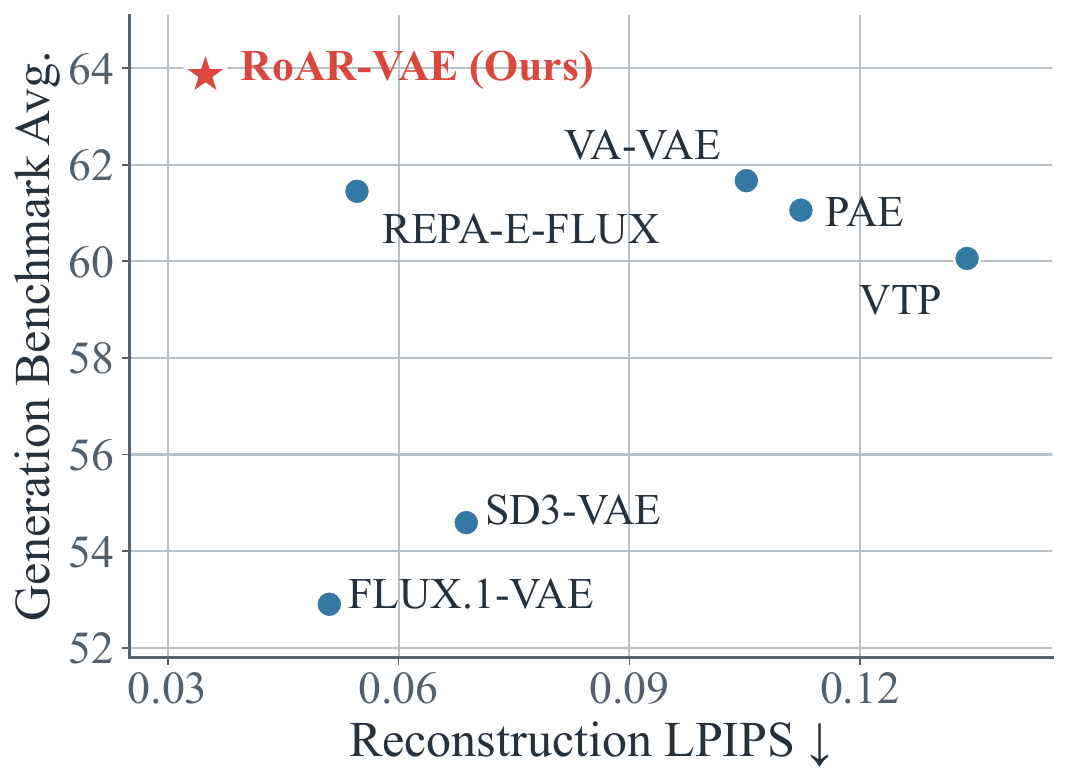}
\vspace{-20pt}
\caption{\textbf{Reconstruction versus generation.} RoAR-VAE achieves a favorable trade-off, combining low reconstruction LPIPS with strong diffusability.}
\label{fig:teaser}
\vspace{-10pt}
\end{wrapfigure}
Empirically, we train RoAR-VAE on ImageNet and evaluate reconstruction and downstream text-to-image generation, using a common diffusion backbone and dataset across tokenizers. RoAR-VAE achieves a favorable balance between reconstruction fidelity and generative learnability (Fig.~\ref{fig:teaser}): it attains the best performance across three text-to-image benchmarks, with an average score 2.19 points above the strongest included baseline. At the same time, it maintains high-fidelity reconstruction, achieving 32.88 dB PSNR and 0.0349 LPIPS at $256^2$ resolution, while improving small-text reconstruction accuracy over FLUX.1-VAE by 16.93 percentage points (Table~\ref{tab:tokbench}). Ablations further show that RobustAlign and RobustRecon provide complementary gains, and that latent corruption is critical to downstream performance. These results support robust prediction as an effective principle for improving latent diffusability while retaining high reconstruction fidelity.


\section{Related Work}
\paragraph{Improving diffusability of tokenizers.} Latent diffusion separates perceptual compression from generative modeling~\citep{ldm,sd3}, motivating visual tokenizers that balance reconstruction fidelity with downstream learnability~\citep{skorokhodov2025improving,yue2026insighttok}.
Pre-trained visual representations~\citep{dinov2} have improved latent diffusion through representation alignment in the denoiser~\citep{repa,irepa} as well as regularization of the tokenizer~\citep{repae,vavae}. Recent work builds tokenizers directly on frozen pre-trained encoders~\citep{rae}, while subsequent approaches address the loss of fine-grained information when repurposing such encoders for generation~\citep{aligntok,psvae,pae,svg}. A complementary line of work jointly optimizes the tokenizer and diffusion model~\citep{lsgm,heek2026unified,unite,repae} to better align latent representations with generative learning.
Other approaches improve tokenizer representations through equivariance and perturbation-based regularization~\citep{eqvae,maetok,detok,ssvae}.

\paragraph{Learning dynamics of diffusion models.}
Training dynamics influence both what diffusion models learn and when. \citet{wang2025an} derive an inverse-variance learning timescale for tractable denoisers, showing that higher-variance covariance modes are learned earlier.

Our work revisits REPA-E~\citep{repae}, a representative approach that combines semantic enhancement with joint tokenizer–diffusion training, and highlights robust prediction as an effective source of supervision for improving latent learnability. The spectral bias of diffusion dynamics offers a complementary explanation: robust prediction makes generation-relevant information more accessible in dominant latent modes, which downstream diffusion models tend to learn earlier.

\section{Method}
\label{sec:method}
\textbf{Overview.} Our goal is to learn a VAE latent space that supports both faithful reconstruction and efficient downstream diffusion learning.
We begin by revisiting REPA-E (\S~\ref{sec:preliminaries}) and isolate the supervision that directly shapes the tokenizer encoder.
This leads to \emph{RobustAlign} (\S~\ref{sec:robust_align}), which predicts pretrained semantic features from corrupted latents without jointly training a full diffusion model.
We then introduce \emph{RobustRecon} (\S~\ref{sec:roar}), which instead reconstructs the clean image from corrupted latents to preserve fine-grained spatial and appearance details that semantic features may underrepresent. Combining the two objectives yields \emph{RoAR-VAE}. Finally, we discuss a possible explanation for the diffusability gains by making semantic and spatial information more accessible in dominant latent modes (\S~\ref{sec:understanding}).

\subsection{Preliminaries}
\label{sec:preliminaries}
\paragraph{Visual tokenizer.}
Given an image $\bm x$, the VAE encoder $E_\phi$ produces a latent representation $\bm r=E_\phi(\bm x)$, from which the decoder $D_\psi$ reconstructs $\bm x$. The tokenizer is trained with the standard VAE objective $\mathcal L_{\mathrm{VAE}}$, which combines pixel reconstruction, perceptual, posterior regularization, and adversarial losses. Latent diffusion operates on the normalized representation $\bm z=\operatorname{Norm}(\bm r)$, where $\operatorname{Norm}(\cdot)$ normalizes the latent statistics.

\paragraph{End-to-end representation alignment (REPA-E).} REPA-E~\citep{repae} jointly trains the autoencoder and diffusion model, as illustrated in Figure~\ref{fig:framework}(a). Let $V_\theta$ denote the diffusion Transformer~\citep{sit,dit}. Under the flow-matching formulation, the noisy latent and diffusion objective are
\begin{equation}
\label{eq:fm}
\begin{aligned}
    \bm{z}_t&=(1-t)\bm{z}+t\bm{\epsilon},
    \quad
    \bm{\epsilon}\sim\mathcal{N}(0,I),
    \quad
    t\sim\mathcal{U}(0,1), \\
    \mathcal{L}_{\mathrm{FM}}(\theta)
    &=
    \left\|
    V_\theta(\operatorname{sg}(\bm{z}_t),t)
    -
    (\bm{\epsilon}-\operatorname{sg}(\bm{z}))
    \right\|_2^2,
\end{aligned}
\end{equation}
where $\bm{z}_t$ denotes the corrupted latent and
$\operatorname{sg}(\cdot)$ denotes the stop-gradient operator.
REPA-E augments this generative objective with representation alignment~\citep{repa}, encouraging intermediate denoiser features to match pretrained visual representations, such as DINO features~\citep{dinov2}:
\begin{equation}
\label{eq:alignment_distance}
\begin{aligned}
\bm z_t(\phi)&=(1-t)\cdot \operatorname{Norm}(E_\phi(\bm x))+t\bm \epsilon,\\
\mathcal L_{\mathrm{REPA}}(\phi,\theta)
&=\frac{1}{N}\sum_{n=1}^{N}
\left(1-\operatorname{Sim}\left(\bm h^{(\ell)}_\theta(\bm z_t(\phi),t)^{[n]},T(\bm x)^{[n]}\right)\right).\\
\end{aligned}
\end{equation}
Here, $\bm z_t(\phi)$ denotes the corrupted latent as a differentiable function of the encoder parameters $\phi$. The feature map $\bm h_\theta^{(\ell)}\in\mathbb R^{N\times d}$ denotes the MLP-projected intermediate features from the $\ell$-th layer of the diffusion Transformer, $n$ indexes the $N$ image patches, $T(\bm x)$ denotes the pretrained visual target, and $\operatorname{Sim}(\cdot,\cdot)$ is cosine similarity. The alignment loss therefore supervises both the diffusion parameters $\theta$ and the tokenizer encoder parameters $\phi$, while the standard VAE objective $\mathcal L_{\mathrm{VAE}}$ is retained throughout training.

To account for changing latent statistics during joint training, REPA-E applies channelwise batch normalization without affine parameters. It tracks the running latent mean $\bm\mu$ and variance $\bm\sigma^2$ and normalizes $\bm z=\operatorname{Norm}(\bm r)=(\bm r-\bm\mu)/{\bm s},\,\bm s=\sqrt{\bm \sigma^2+\delta}$, where $\delta>0$ ensures numerical stability.

A key detail of REPA-E is that the flow-matching loss does \emph{not} backpropagate through the VAE encoder, as such gradients were found to cause latent variance collapse~\citep{repae}.
Specifically, the latent input to $\mathcal L_{\mathrm{FM}}$ is gradient-detached, whereas $\mathcal L_{\mathrm{REPA}}$ remains differentiable through $\bm z_t$ to the encoder. Consequently, among the two diffusion-side objectives, only $\mathcal L_{\mathrm{REPA}}$ provides additional supervision to the VAE encoder.

\subsection{Robust Alignment}
\label{sec:robust_align}

\begin{table}[t]
    \centering
    \caption{\textbf{Controlled comparison with REPA-E.} We train Vanilla-KL-VAE, REPA-E-Tokenizer, and RobustAlign-VAE using the same VAE architecture. For each tokenizer, we train a separate downstream text-to-image diffusion Transformer using the same backbone and training data, and report GenEval~\citep{ghosh2023geneval} performance after 4k, 8k, 16k, and 32k downstream optimization steps. Both REPA-E and RobustAlign substantially improve over the reconstruction-only Vanilla-KL baseline. Results are reported as mean $\pm$ standard deviation over four sampling seeds.}
    \label{tab:repae}
    \resizebox{0.95\linewidth}{!}{
    \begin{tabular}{l|cccc}
        \toprule
        Method & 4k & 8k & 16k & 32k \\
        \midrule
        T2I modeling w. Vanilla-KL-VAE & 11.74$\pm$0.30 & 23.34$\pm$0.57 & 35.47$\pm$0.53 & 47.60$\pm$0.31 \\
        T2I modeling w. REPA-E-Tokenizer & 26.33$\pm$0.25 & 36.42$\pm$0.56 & 49.46$\pm$0.31 & 60.60$\pm$0.90 \\
        T2I modeling w. RobustAlign-VAE
        & 24.56$\pm$0.27 & 39.47$\pm$0.70 & 51.12$\pm$0.35 & 60.89$\pm$0.46 \\
        \bottomrule
    \end{tabular}
    }
\end{table}

\begin{figure}[t]
    \centering
    \includegraphics[width=0.95\linewidth]{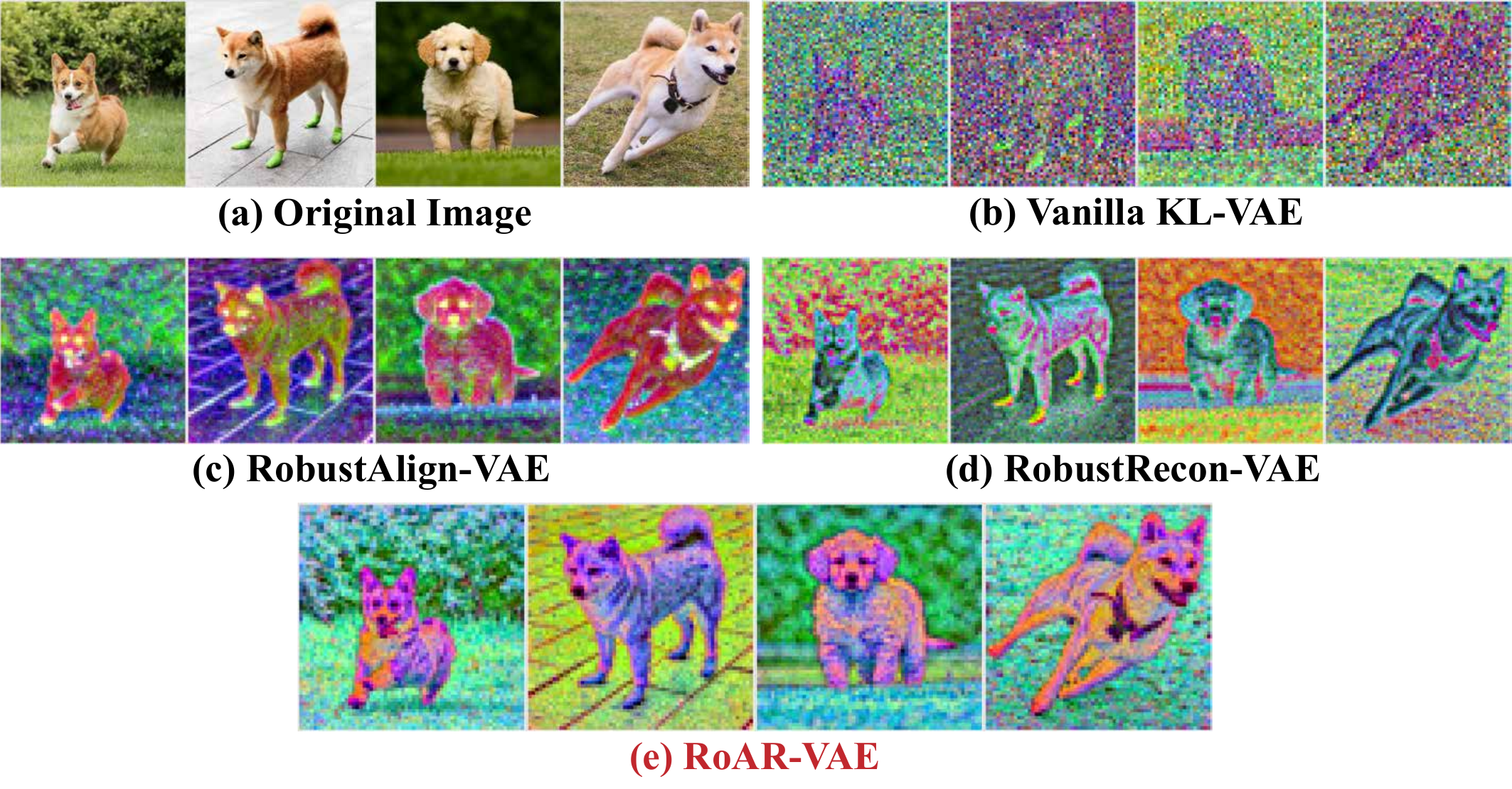}
    \caption{\textbf{PCA visualization of latent representations.} For the input images in (a), we visualize the top three principal components of normalized spatial latent features from (b) KL-VAE, (c) RobustAlign-VAE, (d) RobustRecon-VAE, and (e) RoAR-VAE. RobustAlign emphasizes object-level organization, RobustRecon preserves richer local variation, and RoAR combines both properties. PCA is fitted jointly across the four images for each tokenizer.}
    \label{fig:pca}
\end{figure}

\paragraph{Simplifying REPA-E.}
REPA-E improves the VAE while jointly training a diffusion model, even though the flow-matching loss itself does not update the VAE encoder.
This joint training requires a full generator and also prevents caching latents for large-scale training. We therefore ask: \emph{is the improvement in diffusability driven by joint VAE--diffusion training, or by semantic prediction from corrupted VAE latents?}
To isolate these factors, we introduce \emph{RobustAlign}, shown in Figure~\ref{fig:framework}(b).
RobustAlign-VAE removes the flow-matching objective and all network components used only for diffusion prediction, retaining only the standard VAE objective together with semantic prediction from corrupted latents. Specifically, an auxiliary semantic predictor $F_\eta$ recovers pretrained teacher features $T(\bm x)$ from $\bm z_t$:
\begin{equation}
\label{eq:robust_align}
\mathcal L_{\mathrm{RobustAlign}}(\phi,\eta)
=\frac{1}{N}\sum_{n=1}^{N}
\left(1-\operatorname{Sim}\left(F_\eta(\bm z_t(\phi))^{[n]},T(\bm x)^{[n]}\right)\right).
\end{equation}
The resulting RobustAlign-VAE is trained with
\begin{equation}
\label{eq:robust_align_vae}
    \mathcal{L}_{\mathrm{RobustAlign\text{-}VAE}}
    =
    \mathcal{L}_{\mathrm{VAE}}
    +
    \lambda_{\mathrm{A}}
    \mathcal{L}_{\mathrm{RobustAlign}}.
\end{equation}
Table~\ref{tab:repae} compares the resulting tokenizer with REPA-E under matched downstream diffusion training\footnote{To isolate latent learnability, we train a separate text-to-image diffusion model using the tokenizer produced by REPA-E, without applying the REPA objective during downstream training.}. RobustAlign-VAE performs comparably to REPA-E, indicating that semantic prediction from corrupted latents can recover the diffusability benefit without jointly training the full diffusion model.

\paragraph{Representation view.}
To examine how noisy semantic prediction changes the latent representation, Figure~\ref{fig:pca} visualizes the leading principal components of the latent space. Compared with the reconstruction-only baseline in Figure~\ref{fig:pca}(b), RobustAlign in Figure~\ref{fig:pca}(c) exhibits clearer object-level organization, with more coherent within-object regions and stronger foreground--background separation. This qualitative pattern is consistent with semantic prediction making target-relevant structure more accessible in dominant latent directions.

\subsection{Robust Reconstruction and the RoAR Framework}
\label{sec:roar}
RobustAlign encourages semantically organized latent representations, but semantic targets may underrepresent fine-grained spatial and appearance details that are also important for downstream generation. As illustrated in Fig.~\ref{fig:pca}(c), RobustAlign produces coherent object-level structure, yet local variations within an object, such as texture and boundaries, are less prominent. This motivates a complementary objective that encourages detailed image information to remain recoverable from corrupted latents.

\paragraph{RobustRecon: Preserving fine-grained details.} As shown in Figure~\ref{fig:framework}(c), \textit{RobustRecon} introduces an auxiliary decoder $G_\xi$ that predicts the clean image $\bm x$ directly from the corrupted latent $\bm z_t$:
\begin{equation}
\label{eq:robust_recon}
\mathcal{L}_{\mathrm{RobustRecon}}(\phi,\xi)
=\mathcal L_{\mathrm{MAE}}\left(G_\xi(\bm z_t(\phi)), \bm x\right) + \mathcal L_{\mathrm{LPIPS}}\left(G_\xi(\bm z_t(\phi)), \bm x\right),
\end{equation}
where the two terms measure pixel and perceptual reconstruction error, respectively~\citep{lpips}. $G_\xi$ shares the architecture of the VAE decoder but uses independent parameters and is discarded after tokenizer training. Beyond standard posterior sampling in VAEs~\citep{vae}, RobustRecon applies corruption over a broad range of noise levels, including latents with very low signal-to-noise ratios. The resulting objective backpropagates through $\bm z_t$ to the encoder, encouraging fine-grained image information to remain accessible under corruption. As shown in Fig.~\ref{fig:pca}(d), RobustRecon exposes richer local spatial and appearance variations than RobustAlign, although its object-level foreground--background organization is less pronounced.

\paragraph{Overall framework.} 
RobustAlign encourages noisy latents to preserve high-level semantic structure, whereas RobustRecon encourages them to retain fine-grained image details.
We therefore combine the two objectives into \textit{RoAR-VAE}, short for \textbf{Ro}bust \textbf{A}lignment and \textbf{R}econstruction VAE, as illustrated in Fig.~\ref{fig:framework}(c):
\begin{equation}
\label{eq:roar_objective}
\mathcal L_{\mathrm{RoAR}}=\mathcal{L}_{\mathrm{VAE}}+\lambda_{\mathrm{A}}\mathcal{L}_{\mathrm{RobustAlign}}
+\lambda_{\mathrm{R}}\mathcal{L}_{\mathrm{RobustRecon}},
\end{equation}
where $\lambda_{\mathrm{A}}$ and $\lambda_{\mathrm{R}}$ balance the two auxiliary objectives. Both objectives supervise the encoder through corrupted latents, but use complementary clean targets: pretrained semantic features for RobustAlign and the original image for RobustRecon.
As shown in Figure~\ref{fig:pca}(e), their combination yields latent representations that retain both coherent object-level organization and richer local structure, including clearer boundaries, textures, and intra-object variations.

\subsection{Discussion}
\label{sec:understanding}
\paragraph{Robust prediction promotes key information in dominant modes.}
Prior work suggests that higher-variance latent modes are learned more rapidly by diffusion models~\citep{wang2025an}, indicating that diffusability depends not only on \emph{what} information is encoded, but also on \emph{how} prominently it is represented.
Both RobustAlign and RobustRecon follow the same robust prediction principle: clean targets must remain predictable from corrupted latents. 
Let $\nu_i$ denote the variance of the normalized clean latent $\bm{z}$ along its $i$-th unit covariance eigenvector.
Under the isotropic Gaussian corruption in~\eqref{eq:fm}, the signal and noise variances along this direction are $(1-t)^2\nu_i$ and $t^2$, respectively, yielding
\begin{equation}
\label{eq:directional_snr}
    \operatorname{SNR}_i(t)
    =
    \frac{(1-t)^2\nu_i}{t^2},
    \qquad 0<t<1.
\end{equation}
Thus, at a fixed corruption level, higher-variance modes retain more signal relative to the injected noise~\citep{ssvae}. Optimizing robust prediction thus encourages target-relevant semantics and spatial details to remain accessible in these dominant modes. This provides a possible explanation for the improved diffusability of the resulting latent representations.

\section{Experiments}

\subsection{Experimental Setup}
\label{sec:exp_setup}
We briefly summarize the training and evaluation setup here; full implementation details are provided in Appendix~\ref{app:implementation}.

\textbf{Tokenizer training.}
RoAR-VAE uses a convolutional architecture~\citep{vqgan,ldm} with $8\times$ spatial downsampling, 32 latent channels, and approximately 84M parameters. For controlled comparison with prior tokenizer studies, we train exclusively on ImageNet~\citep{deng2009imagenet} at $256^2$ resolution. Training proceeds in two stages: (1) a 200k-step VAE warm-up using $\mathcal L_{\mathrm{VAE}}$, followed by (2) 250k steps of RoAR refinement with $\lambda_A\!=\!1.5$ and $\lambda_R\!=\!0.5$. For RobustAlign, the semantic predictor $F_\eta$ comprises an 8-layer Transformer~\citep{sit} followed by an MLP projector. It is trained to predict the output features of a frozen DINOv2-B encoder~\citep{dinov2}. For RobustRecon, the pixel predictor $G_\xi$ is initialized from the stage-one VAE decoder $D_\psi$. The corruption level is sampled uniformly as $t\sim\mathcal U(0,1)$.

\textbf{Downstream generative modeling.}
To evaluate latent diffusability, we freeze each tokenizer and train the same text-conditioned diffusion Transformer from scratch on a filtered and recaptioned LAION dataset~\citep{laion} at $256^2$ resolution. We use an MMDiT backbone~\citep{sd3,lens} with 24 layers, a hidden dimension of 1024, and approximately 0.8B parameters, together with a frozen Qwen3-0.6B text encoder~\citep{qwen3}. For tokenizers with different downsampling factors, we adjust the patch size to keep the diffusion sequence length fixed.

\textbf{Baselines.}
We compare against tokenizers from large-scale generative systems, including SD3-VAE~\citep{sd3} and FLUX.1-VAE~\citep{flux2024}, as well as recent methods targeting improved latent diffusability, including VA-VAE~\citep{vavae}, VTP~\citep{vtp}, PAE~\citep{pae}, and REPA-E-FLUX~\citep{repaet2i2025}. We additionally report FLUX.2-VAE~\citep{flux-2-2025} as a reference, but exclude it from baseline comparisons because its tokenizer training data and training recipe are undisclosed.

\textbf{Evaluation protocol.}
For reconstruction, we report rFID, PSNR, LPIPS, and SSIM on the ImageNet validation set at $256^2$ resolution, and evaluate text and facial detail preservation using TokBench~\citep{wu2025tokbench}. For generation, we train a downstream diffusion model for each tokenizer under the controlled setup above and evaluate on GenEval~\citep{ghosh2023geneval}, GenEval2~\citep{kamath2025geneval}, and DPG-Bench~\citep{hu2024ella}. Each score is averaged over four independent sampling seeds, and we report the arithmetic mean across the three benchmarks as a summary metric.

\definecolor{darkgray}{gray}{0.5}
\begin{table}[t]
    \centering
    \caption{\textbf{Reconstruction and downstream generation results.} We report reconstruction metrics evaluated on the ImageNet validation set at $256^2$ resolution. Generation performance is evaluated at 32k downstream updates across three benchmarks: GenEval, GenEval2, and DPG-Bench. Avg. denotes the unweighted mean of the three generation scores.
    }
    \label{tab:main}
    \resizebox{1.0\linewidth}{!}{
    \setlength{\tabcolsep}{4pt}
    \begin{tabular}{l|cccc|cccc}
    \toprule
    \multirow{2}{*}{\textbf{Method}} & \multicolumn{4}{c}{\textbf{Reconstruction}} &  \multicolumn{4}{|c}{\textbf{Generation}} \\ 
     & rFID$\downarrow$ & PSNR$\uparrow$ & LPIPS$\downarrow$ & SSIM$\uparrow$ & GenEval & GenEval2 & DPG-Bench & Avg. \\
    \midrule
    SD3-VAE & 0.2038 & 29.58& 0.0688& 0.8565 & 52.46 &38.03 &73.27 & 54.59\\
    FLUX.1-VAE & 0.1711 & 31.08& 0.0510& 0.8938& 51.79&34.82&72.08 & 52.90\\
    \midrule
    VA-VAE & 0.2892 & 26.26 & 0.1052 & 0.7625& 57.43 & \underline{51.28} & 76.30 & \underline{61.67} \\
    VTP & 0.3457 & 25.75 & 0.1339 & 0.7383& 53.79 & 50.31 & 76.08 & 60.06 \\
    PAE & 0.2643 & 25.90 & 0.1123 & 0.7487& 56.60&49.77&\underline{76.82} & 61.06 \\
    REPA-E-FLUX & 0.1563 & 29.89& 0.0546 & 0.8737& \underline{58.92} & 48.63 & 76.81& 61.45 \\
    \midrule
    RoAR-VAE (Ours) & \textbf{0.1309} & \textbf{32.88}& \textbf{0.0349} & \textbf{0.9258} & \textbf{62.22} & \textbf{51.79}&\textbf{77.56} & \textbf{63.86} \\
    \midrule
    \textcolor{darkgray}{FLUX.2-VAE} & \textcolor{darkgray}{0.1085} & \textcolor{darkgray}{31.52}& \textcolor{darkgray}{0.0398} & \textcolor{darkgray}{0.9052} & \textcolor{darkgray}{60.04} & \textcolor{darkgray}{51.28} & \textcolor{darkgray}{76.88} & \textcolor{darkgray}{62.73} \\
        \bottomrule
    \end{tabular}
    }
\end{table}

\begin{figure}[t]
    \centering
    \includegraphics[width=1.0\linewidth]{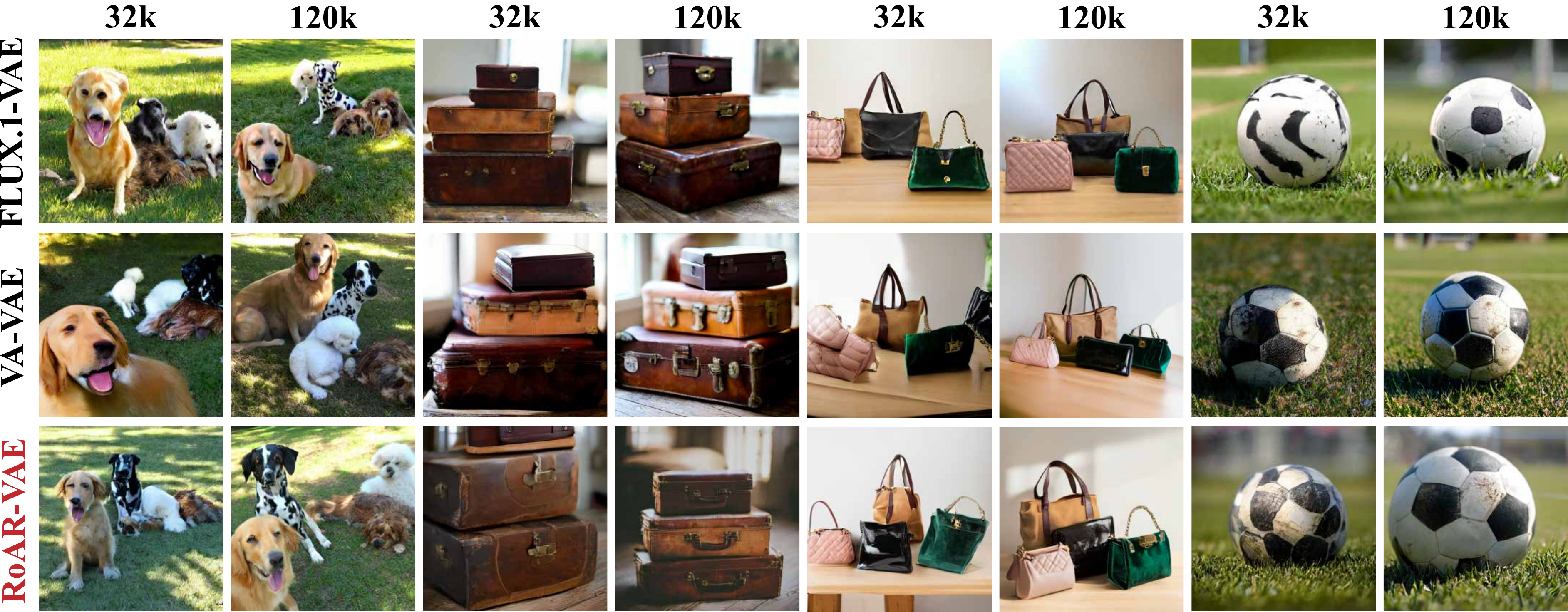}
    \caption{\textbf{Qualitative comparison of text-to-image generation.} Rows compare FLUX.1-VAE, VA-VAE, and RoAR-VAE tokenizers; each subject is shown after 32k and 120k downstream updates.}
    \label{fig:gen_vis}
\end{figure}

\begin{table}[t]
    \centering
    \caption{\textbf{Text and face reconstruction on TokBench,} using text accuracy (T-ACC),
text normalized edit distance (T-NED), and face similarity (F-Sim), evaluated at an image resolution of $256^2$.
    Subscripts $s$ and $m$ denote metrics averaged over small-scale and all instances, respectively.
    Higher is better for every column. Bold denotes the best displayed value.
    }
    \label{tab:tokbench}
    \resizebox{0.85\linewidth}{!}{
    \begin{tabular}{l|cccc|cc}
    \toprule
    \multirow{2}{*}{\textbf{Method}} & \multicolumn{4}{|c|}{\textbf{Text (\%) $\uparrow$}} & \multicolumn{2}{|c}{\textbf{Face $\uparrow$}} \\
    & T-ACC$_s$ & T-ACC$_m$ &  T-NED$_s$ & T-NED$_m$ & F-Sim$_s$ & F-Sim$_m$ \\
    \midrule
    SD3-VAE & 37.29& 68.32& 59.86& 81.36& 0.43 & 0.71\\
    FLUX.1-VAE & 52.07& 76.57& 71.41& 86.96& 0.53& 0.77\\
    VA-VAE & 7.05& 37.15& 25.12& 56.30& 0.22& 0.49\\
    VTP & 0.14& 6.88& 5.87& 20.03& 0.10& 0.24\\
    PAE & 3.31&28.97 & 16.73& 46.97&0.19 & 0.45\\
    \midrule
    RoAR-VAE (Ours) & \textbf{69.00}& \textbf{84.95}& \textbf{82.87}& \textbf{92.22}& \textbf{0.65}& \textbf{0.83}\\
    \midrule
    \textcolor{darkgray}{FLUX.2-VAE} & \textcolor{darkgray}{67.33} & \textcolor{darkgray}{83.80}& \textcolor{darkgray}{81.90}& \textcolor{darkgray}{91.54}&\textcolor{darkgray}{0.63} & \textcolor{darkgray}{0.83} \\
        \bottomrule
    \end{tabular}
    }
\end{table}

\begin{figure}[t]
    \centering
    \includegraphics[width=1.0\linewidth]{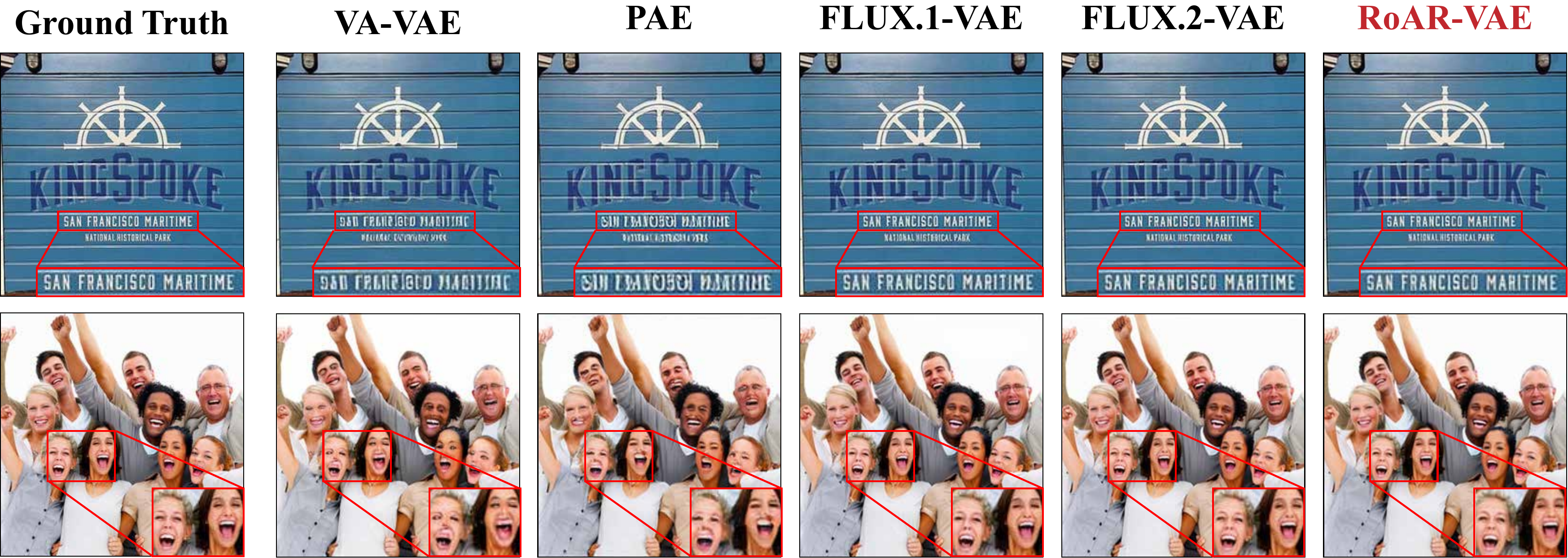}
    \caption{\textbf{Image reconstruction visualization.} Columns show the original image and reconstructions from baseline tokenizers and RoAR-VAE. Images are selected from TokBench~\citep{wu2025tokbench}.}
    \label{fig:recon_vis}
\end{figure}

\subsection{Main Results}

\textbf{Reconstruction versus diffusion learnability.}
Table~\ref{tab:main} compares RoAR-VAE with existing visual tokenizers in terms of reconstruction quality and downstream generation under a fixed 32k-step training budget. RoAR-VAE achieves a favorable balance between the two objectives, substantially improving reconstruction while maintaining strong downstream learnability. Compared with the reconstruction-oriented FLUX.1-VAE, RoAR-VAE improves PSNR by 1.80 dB and reduces LPIPS from 0.0510 to 0.0349, while outperforming the strongest competing learnability-enhanced tokenizer by 2.19 points on the average downstream benchmark score. RoAR-VAE also approaches the reconstruction and generation performance of FLUX.2-VAE, a strong industrial tokenizer with a similarly balanced reconstruction--generation trade-off.

\textbf{Generated image quality.}
Figure~\ref{fig:gen_vis} qualitatively compares downstream generation across different tokenizers. RoAR-VAE produces clearer object shapes and finer visual structure, including more distinct animal contours, sharper suitcase and bag textures, and more regular soccer-ball patterns. Compared with the reconstruction-driven FLUX.1-VAE, these results indicate faster acquisition of object appearance and structure under the same downstream training budget. VA-VAE also exhibits strong learnability, but its generations are noticeably smoother and contain fewer details, consistent with its weaker reconstruction performance relative to RoAR-VAE.

\textbf{Fine-grained reconstruction.}
Table~\ref{tab:tokbench} further evaluates preservation of challenging text and facial details. RoAR-VAE achieves the best performance across all reported text and face reconstruction metrics. Despite being trained from scratch solely on ImageNet, it improves small-text accuracy over FLUX.1-VAE by 16.93 percentage points and increases small-face similarity from 0.53 to 0.65. Although VA-VAE, VTP, and PAE achieve competitive generation scores in Table~\ref{tab:main}, their weaker text and face reconstruction results reveal a substantial loss of fine-grained information. Figure~\ref{fig:recon_vis} provides qualitative examples of these differences, highlighting distortions in textual and facial patterns introduced by the compared tokenizers.

\subsection{Analysis}

\newcommand{\cmark}{\ding{51}} 
\newcommand{\xmark}{\ding{55}} 
\newcommand{\markone}{\ding{182}} 
\newcommand{\marktwo}{\ding{183}} 
\newcommand{\markthree}{\ding{184}} 
\newcommand{\markfour}{\ding{185}} 
\newcommand{\markfive}{\ding{186}} 
\definecolor{darkgreen}{RGB}{0,120,0}
\definecolor{darkred}{RGB}{150,0,0}
\begin{table}[t]
    \centering
    \caption{\textbf{Ablation of robust prediction objectives and latent corruption.} ``Noisy Latent'' indicates whether the auxiliary predictors receive noise-corrupted latents as input.
    Scores are measured after 32k downstream updates; Avg. denotes the mean over the three benchmarks.}
    \label{tab:main_abl}
    \resizebox{0.95\linewidth}{!}{%
    \begin{tabular}{cccc|llll}
    \toprule
     & Noisy Latent & $\mathcal L_{\mathrm{RobustAlign}}$ & $\mathcal L_{\mathrm{RobustRecon}}$ & GenEval & GenEval2 & DPG-Bench & Avg. \\
    \midrule
   \markone & \xmark & \xmark & \xmark & 47.60 & 33.68 & 70.22 & 50.50 \\
  \marktwo &\cmark & \xmark & \cmark & 57.63 & 45.20 & 75.32 & 59.38 \textcolor{darkgreen}{($\uparrow$ 8.88)} \\
  \markthree &\cmark &  \cmark & \xmark & 60.89 & 51.11 & 75.48 & 62.49 \textcolor{darkgreen}{($\uparrow$ 11.99)}\\
  \markfour &\cmark & \cmark & \cmark & \textbf{62.22} & \textbf{51.79} & \textbf{77.56} & \textbf{63.86} \textcolor{darkgreen}{($\uparrow$ 13.36)} \\
  \midrule
  \markfive & \xmark & \cmark & \cmark & 57.54 & 44.01 & 73.86 & 58.47 \textcolor{darkred}{($\downarrow$ 5.39)}\\
    \bottomrule
    \end{tabular}}
\end{table}

\begin{figure}[t]
    \centering
    \includegraphics[width=0.9\linewidth]{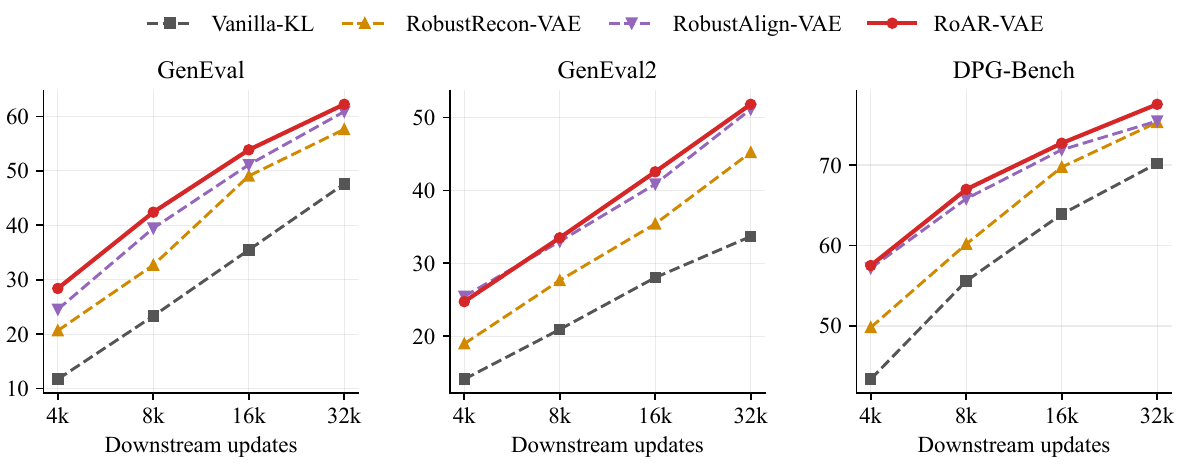}
    \caption{\textbf{Downstream learning curves.} RoAR-VAE consistently outperforms using RobustAlign or RobustRecon alone across different training budgets.
    }
    \label{fig:per_step}
\end{figure}


\textbf{Contribution of the objectives and latent corruption.}
Table~\ref{tab:main_abl} isolates the contributions of the two robust prediction objectives.
Both RobustRecon (\marktwo) and RobustAlign (\markthree) consistently improve over Vanilla-KL (\markone) across all three benchmarks, increasing the average score from 50.50 to 59.38 and 62.49, respectively.
Combining the two objectives (\markfour) further improves performance, reaching an average score of 63.86 and demonstrating the complementary benefits of semantic- and pixel-level robust prediction.
Removing latent corruption (\markfive) substantially degrades the average score from 63.86 to 58.47, highlighting prediction from corrupted latents as a key component of the framework.
Figure~\ref{fig:per_step} further compares downstream performance throughout training: RoAR-VAE consistently outperforms either robust prediction objective used alone across different training budgets.

\begin{wraptable}{r}{0.42\textwidth}
    \centering
    \caption{\textbf{Predictor architecture choices.}}
    \label{tab:choices}
    \vspace{-5pt}
    \resizebox{\linewidth}{!}{%
    \begin{tabular}{l|c}
    \toprule
    Variant & Avg. Score \\
    \midrule
    RobustAlign w. simple MLP $F_\eta$ & 52.70 \\ 
    RobustAlign-VAE & \textbf{62.49} \\
    \midrule
    RoAR w. shared $G_\xi$ and $D_\psi$ & 62.13 \\
    RoAR-VAE & \textbf{63.86} \\
    \bottomrule
    \end{tabular}}
    \vspace{-10pt}
\end{wraptable}
\textbf{Predictor design choices.}
Table~\ref{tab:choices} studies architectural choices for the auxiliary predictors.
For RobustAlign, replacing the Transformer-based semantic predictor $F_\eta$ with a simple MLP causes a substantial performance drop, suggesting that a more expressive prediction pathway is beneficial for shaping diffusion-friendly latents. For RobustRecon, sharing parameters between the pixel predictor $G_\xi$ and the clean reconstruction decoder $D_\psi$ also slightly degrades performance, indicating that the two decoders serve distinct roles in the framework.


\section{Conclusion}

In this paper, we present RoAR-VAE, a visual tokenizer built on the principle of \emph{robust prediction}. By revisiting and simplifying REPA-E, we identify prediction of clean targets from corrupted latents as an effective source of supervision for shaping diffusion-friendly representations. 
We realize this principle with RobustAlign for semantic structure and RobustRecon for fine-grained spatial and appearance details.
Our analysis suggests that robust prediction encourages generation-relevant information to remain accessible in prominent latent modes that are more resistant to corruption and tend to be learned earlier by downstream diffusion models. Empirically, the two objectives provide complementary gains, allowing RoAR-VAE to improve downstream generative learnability while maintaining high-fidelity reconstruction. These results highlight robust prediction as a simple and effective principle for designing visual tokenizers that better balance reconstruction and generation.

\subsection*{AI use statement}

We used generative AI tools for several research tasks that fall under the required disclosure categories, including assisting with implementation of the proposed method, cleaning and reformatting datasets and benchmark inputs, translation, providing feedback on research methodology and experimental design, and assisting with the interpretation of experimental results. We also used generative AI tools for recommended disclosure activities, including developing and editing research code, searching for and identifying relevant literature and research materials, summarizing related work, and drafting and editing portions of the manuscript for clarity and readability.

We did not use generative AI tools to generate synthetic datasets, develop the conceptual framework of the work, formulate or refine research hypotheses, formulate mathematical claims, provide ingredients for mathematical proofs, or write proofs. The proof-related categories are not applicable to this work.

All AI-assisted outputs were reviewed by the authors. AI-generated or AI-edited code was inspected, tested, and verified for correctness before use. The authors take full responsibility for the final content of the paper, including its methodology, experiments, analysis, claims, text, code, and other artifacts produced with the assistance of generative AI.

\subsection*{Ethics statement}
This work evaluates reconstruction performance on a published image reconstruction benchmark~\citep{wu2025tokbench} that includes human faces. However, the proposed visual tokenizer is a general-purpose information compressor and is neither trained specifically on human-centric datasets nor optimized for human-related attributes. For downstream visual generation systems built upon our tokenizer, we recommend adopting appropriate safeguards, such as watermarking and monitoring mechanisms, to mitigate potential misuse.



\subsection*{Reproducibility statement}
We provide detailed descriptions of the proposed method, training objectives, network architectures, experimental setup, and hyperparameter configurations in Section~\ref{sec:method} and Appendix~\ref{app:implementation}. The implementation code and trained models will be released.





\bibliography{refs}
\bibliographystyle{iclr2027_conference}

\clearpage
\appendix

\section{Implementation Details}
\label{app:implementation}

\subsection{Tokenizer Training}
\label{app:tokenizer_training}

RoAR-VAE uses the standard convolutional autoencoder architecture adopted by latent diffusion models~\citep{vqgan,ldm}, with an $8\times$ spatial downsampling factor, 32 latent channels, and approximately 84M parameters. The encoder and decoder each contain four resolution stages with channel widths $(128,256,512,512)$ and two residual blocks per stage. All tokenizer training is performed on ImageNet~\citep{deng2009imagenet} at $256\times256$ resolution.

Training proceeds in two stages. We first warm up the tokenizer for 200k optimization steps using the standard VAE objective $\mathcal L_{\mathrm{VAE}}$. The reconstruction objective uses unit weights for the L1 and LPIPS losses, together with a KL regularization weight of $10^{-6}$. Adversarial supervision is provided by a DINOv2-S~\citep{dinov2}-based discriminator. We then refine the tokenizer for an additional 250k steps using the RoAR objectives with $\lambda_{\mathrm A}=1.5$ and $\lambda_{\mathrm R}=0.5$. The corruption level is sampled uniformly as $t\sim\mathcal U(0,1)$.

For RobustAlign, the semantic predictor $F_\eta$ consists of an 8-layer Transformer~\citep{sit} followed by an MLP projector that matches the dimensionality of the target features. We use features from a frozen DINOv2-B~\citep{dinov2} encoder as the semantic prediction target. For RobustRecon, the pixel predictor $G_\xi$ is initialized from the VAE decoder $D_\psi$ obtained after the warm-up stage and predicts the clean image from corrupted latents using L1 and LPIPS losses. 

The complete RoAR framework is optimized with AdamW~\citep{adamw}, using a batch size of 256 and a learning rate of $1\times10^{-5}$.

\subsection{Downstream Generative Modeling}
\label{app:generative_training}

For each tokenizer, we freeze its parameters and train a text-conditioned diffusion Transformer from scratch on a filtered and recaptioned version of LAION~\citep{laion}. All downstream models are trained at $256\times256$ resolution with a global batch size of 1,024. The diffusion backbone is an MMDiT~\citep{sd3,lens} with 24 Transformer layers, a hidden dimension of 1,024, 16 attention heads, and approximately 0.8B parameters. Text conditioning is provided by a frozen Qwen3-0.6B encoder~\citep{qwen3}.
Because the compared tokenizers use different spatial compression factors, we adjust the latent patch size to maintain the same sequence length for the diffusion Transformer. Specifically, we use a $1\times1$ latent patch size for tokenizers with a downsampling factor of 16 and a $2\times2$ patch size for tokenizers with a downsampling factor of 8. This keeps the number of latent tokens processed by the downstream model consistent across tokenizer variants.

For RoAR-VAE, diffusion timesteps are sampled using shifted logit-normal sampling with exponential shifting and $\mu=1.5$. For each baseline tokenizer, we use its official or recommended timestep scheduler. All downstream models are optimized with AdamW~\citep{adamw}, using $\beta_1=0.9$, $\beta_2=0.999$, a constant learning rate of $2\times10^{-4}$, and a caption dropout probability of 0.1. At inference time, we use an Euler sampler with 50 sampling steps and a classifier-free guidance weight of 5.0.

\subsection{Evaluation Details}
\label{app:evaluation_details}

\paragraph{Baselines.}
\label{app:baselines}
SD3-VAE~\citep{sd3} and FLUX.1-VAE~\citep{flux2024} serve as reconstruction-oriented baselines from large-scale text-to-image systems. VA-VAE~\citep{vavae} directly aligns latent representations with pretrained visual features using a margin-based similarity objective, while VTP~\citep{vtp} jointly optimizes reconstruction and visual representation learning. PAE~\citep{pae} builds on pretrained visual encoders and regularizes the latent manifold for improved downstream generation. REPA-E-FLUX~\citep{repaet2i2025} jointly refines the tokenizer and diffusion model through representation alignment; in our evaluation, we use only the resulting tokenizer and retrain the downstream diffusion model from scratch. We additionally include FLUX.2-VAE~\citep{flux-2-2025} as a strong reference, since its tokenizer training data and recipe are undisclosed.

\paragraph{Reconstruction.}
We evaluate reconstruction on the ImageNet~\citep{deng2009imagenet} validation set at $256\times256$ resolution and report reconstruction FID (rFID), PSNR, LPIPS, and SSIM. To further assess preservation of fine-grained content, we evaluate on TokBench~\citep{wu2025tokbench}, which contains challenging text- and face-centric images. For text reconstruction, TokBench reports exact recognition accuracy (T-ACC) and normalized edit distance (T-NED), with separate metrics for small text instances and all text instances. For face reconstruction, it measures cosine similarity between face-recognition embeddings (F-Sim), likewise reporting both small-face and overall results. These metrics complement standard image-level reconstruction measures by directly evaluating whether tokenizers preserve small, semantically important details.

\paragraph{Text-to-image generation.}
For each tokenizer, we train a downstream diffusion model following Appendix~\ref{app:generative_training}. Unless otherwise stated, benchmark results are measured after 32k downstream optimization steps, enabling comparison of latent diffusability under a fixed training budget. We evaluate text--image alignment on three complementary benchmarks. GenEval~\citep{ghosh2023geneval} evaluates compositional generation involving object co-occurrence, spatial relations, counting, and colors, while GenEval2~\citep{kamath2025geneval} extends this setting with broader coverage of visual concepts and higher compositionality. DPG-Bench~\citep{hu2024ella} focuses on dense prompts containing multiple objects, detailed attributes, and complex relationships. Each reported score is averaged over four independent sampling seeds. We additionally report the arithmetic mean of the three benchmark scores, denoted \emph{Avg.}, as a compact summary.

\end{document}